\documentclass[letterpaper, 10 pt, conference]{ieeeconf}  

\IEEEoverridecommandlockouts                              

\usepackage{graphics} 
\usepackage{epsfig} 
\usepackage{mathptmx} 
\usepackage{times} 
\usepackage{amsmath} 
\usepackage{amssymb}  
\usepackage{subfigure}
\usepackage{color}
\graphicspath{ {figs/} }
\usepackage[space]{cite}

 \usepackage[hidelinks]{hyperref}
\usepackage[algoruled]{algorithm2e}
\SetAlFnt{\small}
\usepackage{multirow}
\usepackage{balance}

\title{\LARGE \bf
Contact-Implicit Trajectory Optimization Based on a Variable Smooth Contact Model and Successive Convexification
}

\author{Aykut \"{O}zg\"{u}n \"{O}nol, Philip Long, and Ta\c{s}k\i n Pad\i r$^{1}$
\thanks{This research is supported by the Department of Energy under Award No. DE-EM0004482, by the National Aeronautics and Space Administration under Grant No. NNX16AC48A issued through the Science and Technology Mission Directorate, by the National Science Foundation under Award Nos. 1451427 and 1544895, and by the Office of the Secretary of Defense under Agreement No. W911NF-17-3-0004.}
\thanks{$^{1}$RIVeR Lab, Northeastern University, Boston, MA, USA
        {\tt\small \{onol.a, p.long, t.padir\}@northeastern.edu}}%
}

\begin{document}

\maketitle
\thispagestyle{empty}
\pagestyle{empty}

\begin{abstract}
In this paper, we propose a contact-implicit trajectory optimization (CITO) method based on a variable smooth contact model (VSCM) and successive convexification (SCvx). The VSCM facilitates the convergence of gradient-based optimization without compromising physical fidelity. On the other hand, the proposed SCvx-based approach combines the advantages of direct and shooting methods for CITO. For evaluations, we consider non-prehensile manipulation tasks. The proposed method is compared to a version based on iterative linear quadratic regulator (iLQR) on a planar example. The results demonstrate that both methods can find physically-consistent motions that complete the tasks without a meaningful initial guess owing to the VSCM. The proposed SCvx-based method outperforms the iLQR-based method in terms of convergence, computation time, and the quality of motions found. Finally, the proposed SCvx-based method is tested on a standard robot platform and shown to perform efficiently for a real-world application.
\end{abstract}

\section{Introduction}
Contact-implicit trajectory optimization (CITO) is an attractive approach that enables the planning of contact-rich, complex motions without a predefined contact schedule. This problem is non-convex due to nonlinear dynamics. Thus, generally, it is transcribed into a finite-dimensional (shooting or direct \cite{betts1998survey}) optimization problem and solved by nonlinear programming (NLP) or variants of differential dynamic programming (DDP).

Solving a shooting optimization problem using NLP is usually not feasible. Alternatively, a CITO problem can be posed as a more efficient direct optimization problem \cite{posa2014direct}. However, this requires imposing rigid-body dynamics with collisions and frictional effects as constraints. In addition, generic NLP methods usually have unreliable convergence properties, and the computational resources required to solve the problem are unbounded~\cite{acikmese2018scvx}. On the other hand, variants of DDP such as iterative linear quadratic regulator (iLQR) \cite{todorov2004ilqr} and sequential linear quadratic (SLQ) control \cite{sideris2005slq} seem to be effective for CITO problems. Furthermore, when used with smooth contact models, they can be implemented in real time \cite{todorov2015mpc,buchli17_mpc}. However, it is difficult to include constraints, and they require a feasible initial guess.

Mao et al. \cite{acikmese2016scvx} proposed successive convexification (SCvx) to solve trajectory optimization problems subject to nonlinear dynamics with a global convergence guarantee. In \cite{acikmese2018scvx}, SCvx was extended to handle non-convex constraints and shown to exhibit superlinear convergence and outperform generic sequential quadratic programming. Moreover, this method is suitable for real-time applications since convex subproblems can be solved very quickly \cite{boyd2013ecos,acikmese2017quad,acikmese2019aero}.

At each iteration of SCvx, the nonlinear dynamic constraints are linearized about the solution from the previous succession, and the resulting convex problem is solved for the state and control trajectories simultaneously subject to the linearized constraints, similar to direct optimization. In this study, we modify the SCvx algorithm to combine the benefits of direct and shooting methods. That is, we apply changes only for controls and roll-out the nonlinear dynamics to find state trajectories. As a result, the trajectories are dynamically consistent, as in shooting methods. Moreover, SCvx keeps track of the fidelity of approximations over successions by comparing the cost improvements for linear and nonlinear dynamics, which eliminates the line search step used in the DDP variants. Thus, the proposed SCvx algorithm enjoys the accuracy of shooting methods while still having the numerical advantages of direct methods.

\begin{figure}
  \centering
  \includegraphics[width=.8\columnwidth]{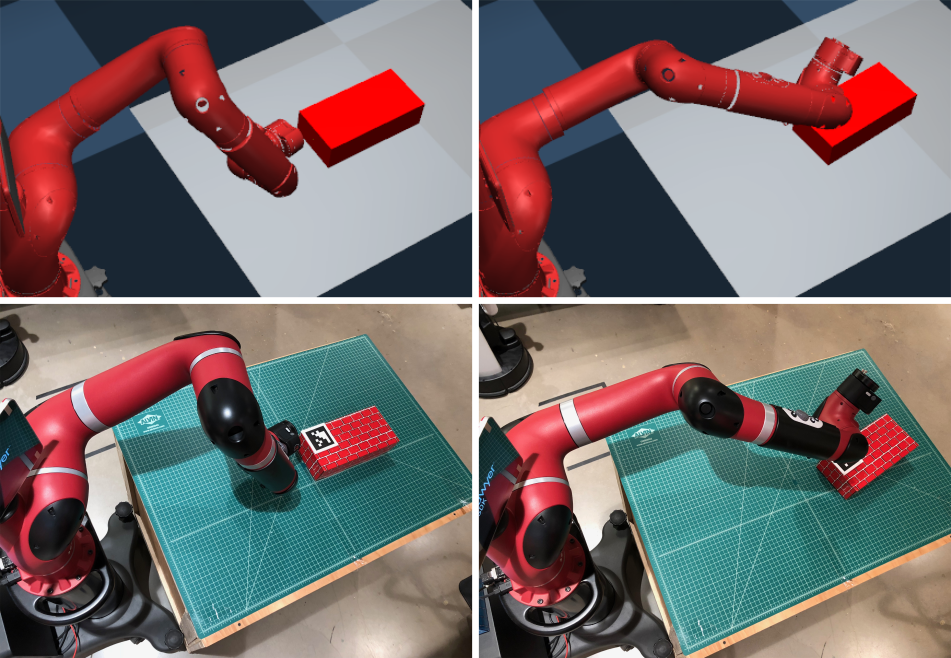}
  \caption{Initial (left) and final (right) snapshots of a 7 DOF arm (Sawyer) performing a pushing task in simulation (top) and experimentally (bottom).}
  \label{fig:sim_vs_exp}
\end{figure}

Another essential factor in CITO is the contact model. While smooth models facilitate convergence, they also lead to physical inaccuracies. In order to address this problem, we recently proposed a variable smooth contact model (VSCM)~\cite{onol2018comparative}. In this approach, smooth virtual forces are exploited to discover contacts, and existing contact mechanics in a physics engine is utilized for the simulation of frictional contacts. The virtual forces are suppressed throughout the optimization. As a result, physically-accurate motions can be obtained while maintaining fast convergence.

In this study, we propose a CITO method based on the VSCM and successive convexification. We compare this approach to an iLQR-based variant in terms of convergence, computation time, and the quality of motions. Although this framework is task independent, we focus on non-prehensile manipulation tasks. For comparisons, we consider a planar 4 degree-of-freedom (DOF) robot pushing a box in different directions. Additionally, the proposed method is demonstrated on a standard 7 DOF robot manipulator by employing a state-of-the-art physics engine and a sparse solver to exploit the structure of the problem. Finally, we execute the planned motions on the hardware to verify the physical fidelity of the planned motions, see Fig. \ref{fig:sim_vs_exp}.

\subsection{Related Work}
The main objective of contact-implicit trajectory optimization is to solve for non-smooth trajectories with impacts and discontinuities as a finite-dimensional optimization problem. For this purpose, Yunt \& Glocker \cite{yunt2005trajectory} proposed transcribing such a problem into an NLP with complementarity constraints and a bi-level optimization where the control inputs are optimized at the higher level while the states and contact forces are optimized at the lower level. On the other hand, Mordatch et al. \cite{todorov2012discovery,todorov2012manipulation} generate complex behaviors for animated characters using quadratic programming with soft constraints. Nevertheless, in these studies, various relaxations are made to solve a non-convex problem efficiently by compromising physical realism.

In order to make CITO numerically more tractable without relaxations, Posa et al. \cite{posa2014direct} proposed a direct method with complementarity constraints based on the time-stepping scheme initially proposed in \cite{stewart1996implicit}. Gabiccini et al. \cite{gabiccini2018computational} use such an approach for grasping and dexterous manipulation with penalty- and complementarity constraints-based contact models tailored for manipulation problems. In order to improve convergence, \cite{manchester2017variational} relaxes the complementarity constraints, while \cite{marcucci2017two} and \cite{mastalli2016hierarchical} propose hierarchical strategies to warm start the optimization subject to strict constraints. Winkler et al. \cite{buchli2018gait} recently proposed another direct method (without complementarity constraints) based on a continuous parameterization of feet motion and forces to generate gait sequences. Direct methods have successfully obtained highly-dynamic, complex motions, yet they are typically not suitable for real-time applications.

On the other hand, the Hessian-approximating DDP variants iLQR \cite{todorov2004ilqr} and SLQ control \cite{sideris2005slq} can solve CITO problems much faster. Tassa et al. \cite{todorov2012synthesis} presented a model predictive control (MPC) method based on iLQR and a smooth fragment of complementarity constraints that runs in simulation in near real time. In \cite{todorov2015mpc}, the method is validated on a humanoid robot hardware for quasi-static motions. Neunert et al. \cite{buchli16_relax} use SLQ control to generate trotting gaits on a quadruped robot. In this work, the complementarity constraints are converted into soft constraints in terms of state and control constraints; yet experimental results prove that resulting motions are physically feasible. In \cite{buchli17_discovery}, highly-dynamic motions for a quadruped are planned within a minute by using a smooth contact model and SLQ control without constraints. Moreover, \cite{buchli17_mpc} showed that such a method can achieve real-time operation in an MPC fashion owing to the smooth contact model. However, there is no guarantee that resulting motions would respect the physical limits, even though they remained feasible in this case. Recently, Carius et al. \cite{hutter2018trajectory} proposed a bi-level optimization approach based on Moreau's time-stepping scheme \cite{moreau1988unilateral} and iLQR such that contact constraints are taken into account in the system dynamics while iLQR solves the unconstrained lower-level problem. The proposed approach is compared to the smooth contact model-based method in \cite{buchli17_discovery,buchli17_mpc}. It is noted that it is difficult to tune the smooth contact model.

In order to combine the advantages of shooting and direct methods, Giftthaler et al. \cite{buchli2017family} proposed a multiple shooting version of iLQR for trajectory optimization. They also consider a CITO problem to generate trotting on a quadruped with a trivial initial guess, where the robot stands still. The results indicate that both the single-shooting and the proposed multiple-shooting iLQR methods converge linearly but the hybrid method converges in fewer iterations. However, this method is limited to unconstrained problems.

\subsection{Contributions}
The contributions of this work are the following:
\begin{itemize}
    \item A contact-implicit trajectory optimization method based on the variable smooth contact model and successive convexification (SCvx-VSCMO) is proposed.
    \item An iLQR-based variant of the proposed approach is implemented for a comparative analysis. The results show that both methods can efficiently find physically-feasible motions that fulfill task requirements with a trivial initial seed, owing to the VSCM; while the SCvx-VSCMO outperforms the iLQR-VSCMO in terms of convergence, computation time, and the quality of motions.
    \item The SCvx-VSCMO is evaluated on a standard robot platform, Sawyer. It is demonstrated that our proposed method can solve a practical problem in an effective manner by utilizing a state-of-the-art physics engine and a sparse solver. Moreover, the physical feasibility of the motions generated by our approach is verified by executing the planned motions on the hardware.
\end{itemize}


\section{Methodology}
\subsection{Dynamic Model}
The dynamics of an $n_r$ DOF robot is given by
\begin{equation}
     \mathbf{M}(\mathbf{q}) \ddot{\mathbf{q}} + \mathbf{c}(\mathbf{q},\dot{\mathbf{q}}) = \boldsymbol{\tau} + \mathbf{J}_{ext}(\mathbf{q})^T \boldsymbol{\lambda}_{ext},
     \label{eq:dynamics}
\end{equation}
where $\mathbf{q}, \dot{\mathbf{q}}, \ddot{\mathbf{q}} \in \mathbb{R}^{n_r} $ are the vectors of joint positions, velocities, and accelerations; $\mathbf{M}(\mathbf{q}) \in \mathbb{R}^{n_r \times n_r}$ is the mass matrix; $\mathbf{c}(\mathbf{q},\dot{\mathbf{q}}) \in \mathbb{R}^{n_r}$ represents the Coriolis, centrifugal, and gravitational terms; $\boldsymbol{\tau} \in \mathbb{R}^{n_r}$ is the vector of generalized joint forces; $\mathbf{J}_{ext}(\mathbf{q}) \in \mathbb{R}^{6n_{ext} \times n_r}$ is the Jacobian matrix mapping the joint velocities to the Cartesian velocities at the external contact points; and $\boldsymbol{\lambda}_{ext} \in \mathbb{R}^{6n_{ext}}$ is the vector of generalized contact forces at the contact points for $n_{ext}$ external contacts.

In this study, the generalized joint forces are decomposed as $\boldsymbol{\tau}=\boldsymbol{\tau}_u+\Tilde{\mathbf{c}}$, where $\Tilde{\mathbf{c}}$ is an estimation of $\mathbf{c}(\mathbf{q},\mathbf{\dot{q}})$ and $\boldsymbol{\tau}_u\in\mathbb{R}^{n_r}$ is the vector of control variables in terms of joint forces, so that this control term is linearly related to the joint accelerations in the absence of external contact. 

The state vectors containing joint positions and velocities for the robot and free bodies (e.g., objects) are denoted respectively by $\mathbf{x}_r \in \mathbb{R}^{2n_r}$ and $\mathbf{x}_f \in \mathbb{R}^{6n_f}$. Hence, the overall state vector $\mathbf{x}\in\mathbb{R}^n$ is defined as $\mathbf{x} = [\mathbf{x}_r^T, \mathbf{x}_f^T]^T$, i.e., $n=2n_r+6n_f$.

\subsection{Contact Model}
We assume that there are $n_c$ contact pairs of predefined contact candidates on the robot and in the environment. For the $i^{th}$ contact pair, the magnitude of the normal contact force $\gamma_i$ is calculated from the signed distance between the contact candidates on the robot and in the environment $\phi_i$, the virtual stiffness $k_i$, and the curvature $\alpha$ by $\gamma_i(\mathbf{x}) = k_i e^{-\alpha \phi_i(\mathbf{x})}$. The corresponding virtual force acting on the free body associated with the contact pair $\boldsymbol{\lambda}_i \in \mathbb{R}^6$ is then calculated by:
\begin{equation}
    \boldsymbol{\lambda}_i(\mathbf{x}) = \gamma_i(\mathbf{x})
    \begin{bmatrix}
    \mathbb{I}_3 \\ \mathbf{\hat{l}}_i    
    \end{bmatrix}
    \mathbf{n}_i(\mathbf{x}),
\end{equation}
where $\mathbb{I}_3$ is $3\times3$ identity matrix; $\mathbf{l}_i$ is the vector from the center of mass of the body to the contact candidate on the robot; $\hat{\mathbf{l}}_i$ is the skew-symmetric matrix form of $\mathbf{l}_i$; and $\mathbf{n}_i \in \mathbb{R}^3$ is the contact surface normal. Then, the net virtual force acting on a free body is the sum of the virtual forces associated with the contact candidates on that body. Consequently, the states of the robot and the free bodies are related through virtual forces.

In the variable smooth contact model, $\mathbf{k}\in\mathbb{R}^{n_c}$ is a part of the control input; namely, $\mathbf{u}\in \mathbb{R}^m$ consists of $\boldsymbol{\tau}_u$ and $\mathbf{k}$, i.e., $\mathbf{u}=[\boldsymbol{\tau}_u^T, \mathbf{k}^T]^T$ and $m=n_r+n_c$. Hence, $\mathbf{k}$ is a decision variable of optimization and initialized with large values such that there is a non-zero virtual force on each contact candidate. Nonetheless, it is also penalized as a cost term, and hence virtual forces vanish as the optimization converges.

In this work, there are several extensions with respect to the VSCM proposed in \cite{onol2018comparative}. First, $\mathbf{k}$ is penalized instead of $\boldsymbol{\gamma}$ because $\mathbf{k}$ is a control input while $\boldsymbol{\gamma}$ is a nonlinear term. Second, $\alpha$ is constant since it has been observed that $\alpha$ being a function of $\mathbf{k}$ does not improve the performance in this case.

\subsection{Trajectory Optimization}

A trajectory optimization problem can be transcribed into a finite-dimensional optimization problem by discretization over a finite time horizon. A nonlinear discrete-time dynamic model can be written as $\mathbf{x}_{i+1} = f(\mathbf{x}_i,\mathbf{u}_i)$ which describes the evolution of states over the time step $i$ given the control input $\mathbf{u}_i$ according to the nonlinear dynamics $f:\mathbb{R}^n\times\mathbb{R}^m\rightarrow\mathbb{R}^n$. In this study, we assume a zero-order hold for the controls over a time step. Then, the finite-dimensional trajectory optimization problem can be written in terms of state and control trajectories $\mathbf{X} \triangleq [\mathbf{x}_1,...,\mathbf{x}_{N+1}]$ and $\mathbf{U} \triangleq [\mathbf{u}_1,...,\mathbf{u}_{N}]$; final and integrated cost terms $C_F$ and $C_I$; lower and upper control and state limits $\mathbf{u}_L$, $\mathbf{u}_U$, $\mathbf{x}_L$, and $\mathbf{x}_U$; and the number of time steps $N$, as follows:
\begin{subequations}
    \begin{align}
        \underset{\mathbf{U}}{\text{minimize }} C(\mathbf{X},\mathbf{U}) \triangleq C_F(\mathbf{x}_{N+1}) + \sum_{i=1}^{N} C_I(\mathbf{x}_i,\mathbf{u}_i)
    \end{align}
subject to:
    \begin{align}
        & \mathbf{x}_{i+1} = f(\mathbf{x}_i,\mathbf{u}_i) \text{ for } i=1,...,N, \label{eq:dynamics_constraint} \\
        & \mathbf{u}_{L} \leq \mathbf{u}_{1,...,N} \leq \mathbf{u}_{U}, \ \mathbf{x}_{L} \leq \mathbf{x}_{1,...,N+1} \leq \mathbf{x}_{U}.
    \end{align}
\end{subequations}
In a shooting method, only the control inputs are the decision variables while \eqref{eq:dynamics_constraint} is enforced; whereas, in a direct method, the states and the control inputs are optimized simultaneously, and \eqref{eq:dynamics_constraint} is a constraint.

Both trajectory optimization methods considered in this work, i.e., SCvx and iLQR, try to solve this problem. Moreover, both methods require the linearization of the system dynamics about the trajectory from the previous iteration $s$ $(\mathbf{X}^s, \mathbf{U}^s)$. The first-order approximation of \eqref{eq:dynamics_constraint} is given by:
\begin{equation}
    \delta \mathbf{x}_{i+1} = \mathbf{A}_i \delta \mathbf{x}_i + \mathbf{B}_i \delta \mathbf{u}_i,
\end{equation}
where $\mathbf{A}_i \triangleq \partial f(\mathbf{x}_i,\mathbf{u}_i) / \partial    \mathbf{x}_i |_{\mathbf{x}_i^s,\mathbf{u}_i^s}$, $\mathbf{B}_i \triangleq \partial f(\mathbf{x}_i,\mathbf{u}_i) / \partial \mathbf{u}_i|_{\mathbf{x}_i^s,\mathbf{u}_i^s}$, $\delta \mathbf{x}_i = \mathbf{x}_i - \mathbf{x}_i^s$, and $\delta \mathbf{u}_i = \mathbf{u}_i - \mathbf{u}_i^s$ for $i=1,...,N$.

\subsection{Successive Convexification}
The successive convexification algorithm is based on repeating the following steps in successions: linearizing the dynamics and nonlinear constraints about the state and control trajectories from the previous succession; solving the resulting convex subproblem subject to linearized dynamic, state, and control constraints within a trust region to avoid artificial unboundedness due to linearization; and updating the radius of the trust region based on the similarity between linear and nonlinear penalized costs. A thorough description of the algorithm and an analysis of its convergence properties are presented in \cite{acikmese2018scvx}.

The convex subproblem that employs unbounded virtual controls $\mathbf{v}\in \mathbb{R}^n$ to prevent artificial infeasibility due to linearization is given by:
\begin{subequations}
    \begin{align}
        &\underset{\mathbf{\delta X},\mathbf{\delta U},\mathbf{V}}{\text{minimize }} L \triangleq C(\mathbf{X}^s + \mathbf{\delta X},\mathbf{U}^s + \mathbf{\delta U}) + \kappa\sum_{i=1}^{N}||\mathbf{v_i}||_1
    \end{align}
subject to:
    \begin{align}
        & \mathbf{\delta x}_{i+1} = \mathbf{A}_i \mathbf{\delta x}_i + \mathbf{B}_i \mathbf{\delta u}_i + \mathbf{v}_i \text{ for } i=1,...,N,  \label{eq:scvx_dynamics_constraint}  \\
        & \mathbf{x}_{L} \leq \mathbf{x}_i^s + \mathbf{\delta x}_i \leq \mathbf{x}_{U} \text{ for } i=1,...,N+1, \\
        & \mathbf{u}_{L} \leq \mathbf{u}_i^s + \mathbf{\delta u}_i \leq \mathbf{u}_{U} \text{ for } i=1,...,N, \\
        & ||\mathbf{\delta X}||_1 + ||\mathbf{\delta U}||_1 \leq r^s,
    \end{align}
\end{subequations}
where $\mathbf{\delta X} \triangleq [\mathbf{\delta x}_1,...,\mathbf{\delta x}_{N+1}]$, $\mathbf{\delta U} \triangleq [\mathbf{\delta u}_1,...,\mathbf{\delta u}_{N}]$, $\mathbf{V} \triangleq [\mathbf{v}_1,...,\mathbf{v}_{N}]$, $\kappa$ is the penalty weight, and $r$ is the trust region radius. This subproblem is a simultaneous (direct) optimization problem. Thus, it has a larger size but a sparse structure, since in \eqref{eq:scvx_dynamics_constraint}, each state depends only on the state and control input of the previous time step. This sparsity can be exploited by using a sparse convex programming solver.

In this study, we make the following modification in the original SCvx algorithm in order to make it more efficient for our application. When the subproblem is solved, we apply only the change of controls and roll-out the nonlinear dynamics, rather than applying the change of states obtained from the subproblem. The reason is that the hard enforcement of the dynamics eliminates the accumulation of defects (i.e., $f(x_i,u_i) - x_{i+1}$) that may occur when using the penalty approach in the original method as well as improves the convergence in our experiments by allowing larger trust regions. Accordingly, we calculate the actual improvement based on the cost value evaluated for the nonlinear trajectory, without the penalty on the defects. The modified procedure of SCvx is summarized in Algorithm \ref{alg:scvx}. We stress the fact that the convergence properties of the original SCvx are not guaranteed for the modified SCvx but it still provides fast and reliable convergence in our experiments.

\begin{algorithm}[!bt]
    \SetAlgoLined
    \SetKwInOut{Input}{Input}
    \Input{Initial state vector $\mathbf{x}_1$, initial control trajectory $\mathbf{U}^1$, initial trust region radius $r^1>0$, $\kappa>0$, $0 < \rho_0 < \rho_1 < \rho_2 < 1$, $r_{max}>0$, $\beta_{shrink}>1$  $\beta_{expand}>1$, $s_{max}>1$, $\Delta L_{tol}>0$.}
    
    Roll-out the dynamics: $\mathbf{x}^{1}_{i+1} \leftarrow f(\mathbf{x}^{1}_i,\mathbf{u}^{1}_i)$ for $i = 1,...,N$. \\
    \Repeat{$s > s_{max}$ or $|\Delta L^s| \leq \Delta L_{tol}$}
    {
        \textbf{Step 1} Linearize the dynamics about $(\mathbf{X}^{s},\mathbf{U}^{s})$. \\
        \textbf{Step 2} Solve the convex subproblem for $(\delta \mathbf{X},\delta \mathbf{U})$. \\
        \textbf{Step 3} Roll-out the dynamics: \\
        $\mathbf{x}_{i+1} \leftarrow f(\mathbf{x}_i,\mathbf{u}^{s}_i+\delta \mathbf{u}_i)$ for $i = 1,...,N$. \\
        \textbf{Step 4} Compute the actual and predicted improvements in the cost: $\Delta C^s \triangleq C(\mathbf{X}^s,\mathbf{U}^s) - C(\mathbf{X},\mathbf{U}^s+\mathbf{\delta U})$ and $\Delta L^s \triangleq C(\mathbf{X}^s,\mathbf{U}^s) - L(\mathbf{X}^s+\mathbf{\delta X},\mathbf{U}^s + \mathbf{\delta U})$. \\
        Compute the similarity ratio $\rho^s \triangleq \Delta C^s/\Delta L^s$. \\
        \eIf{$\rho^s<\rho_0$}
        {
            Reject the solution, shrink the trust region $r^{s} \leftarrow r^s/\beta_{shrink}$, and go to \textbf{step 2}.
        }
        {
            Accept the solution $\mathbf{U}^{s+1} \leftarrow \mathbf{U}^s + \mathbf{\delta U}$, $\mathbf{X}^{s+1} \leftarrow \mathbf{X}$. \\
            Update the trust region radius by
            \begin{equation}
                r^{s+1} = 
                \begin{cases}
                    r^s/\beta_{shrink}, \hspace{0.2cm} \text{if } \rho^s < \rho_1; \\
                    r^s,                \hspace{1.2cm}\text{if } \rho_1 \leq \rho^s < \rho_2; \\
                    r^s\beta_{expand},  \hspace{0.25cm} \text{if } \rho_2 \leq \rho^s.
                \end{cases} \nonumber
            \end{equation}
        }
        $r^{s+1} \leftarrow \text{max}(r^{s+1},r_{max})$, $s \leftarrow s+1$ 
    }
    \KwRet{$(\mathbf{X}^{s},\mathbf{U}^{s})$.}
    \caption{Successive Convexification}
    \label{alg:scvx}
\end{algorithm}

\section{Application: Non-Prehensile Manipulation}
\subsection{Problem Description}
\begin{figure}
  \centering
  \includegraphics[width=.5\columnwidth]{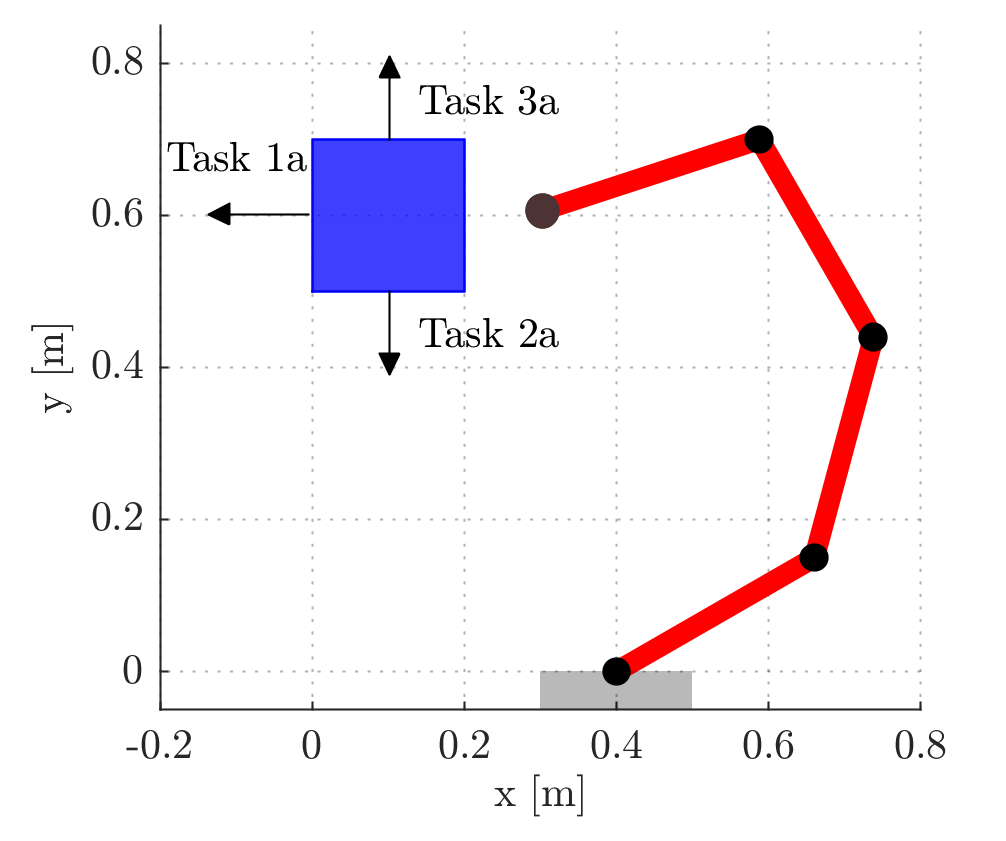}
  \caption{The simulation environment with a 4 DOF planar robot and a box to be manipulated and an illustration of Tasks 1a, 2a, and 3a.}
  \label{fig:matlab_task}
\end{figure}

Although the methodology presented here can be generalized to both locomotion and manipulation tasks, we consider non-prehensile manipulation tasks for evaluations in this study. First, we focus on a planar example in which a planar 4 DOF robot arm manipulates a box, as shown in Fig. \ref{fig:matlab_task}. Here, the end effector and each edge of the object is defined as a contact candidate (i.e., $n_c=4$). We consider three pushing tasks in which the object needs to be moved 10 cm in $-x$, $-y$, and $+y$ directions in 1~s, i.e., Tasks 1a, 2a, and 3a. 

We also demonstrate the applicability and performance of the proposed SCvx-VSCMO method on a standard 7 DOF robot arm, Sawyer from Rethink Robotics, for pushing a box forward a desired distance without rotating it, as depicted in Fig. \ref{fig:sim_vs_exp}. In this case, there is one contact candidate in the environment that is the closest surface of the object to the robot. Two tasks are considered for desired distances of 5 cm and 10 cm, i.e., Tasks 1b and 2b. The planned motions are executed on the hardware as well to show that they are physically feasible and similar behaviors can be obtained in experiments. Motions found for both planar and 7 DOF arm applications are demonstrated in the accompanying video.

\subsection{Iterative Linear Quadratic Regulator}


We use iLQR both for comparisons and to show that our variable smooth contact model can also be used with a DDP variant. A control-limited version of iLQR is proposed in \cite{todorov2014control} which solves a quadratic problem in the backward pass to handle box constraints on controls. Since the robots considered in this study have torque constraints, we employ this version of iLQR.

\subsection{Cost Function and Constraints}
Manipulation tasks can be specified based on desired object configurations. In this study, the cost function is written in terms of a quadratic final cost based on the deviations of the position and orientation of the object from a desired pose, $\mathbf{p}_o^e$ and $\boldsymbol{\theta}_o^e$:
\begin{equation}
    C_F = w_1 ||\mathbf{p}_o^e||_2^2 + w_2 ||\boldsymbol{\theta}_o^e||_2^2, \label{eq:cf}
\end{equation}
and a quadratic integrated cost that penalizes the control variable $\mathbf{k}$, and hence the virtual forces:
\begin{equation}
    C_I = w_3 ||\mathbf{k}_i||_2^2,                                         \label{eq:ci}
\end{equation}
where $w_1$, $w_2$, and $w_3$ are the weights.

For the planar example, the weights are tuned by considering the individual performance of each method. They are selected as $w_1=10^4$ and $w_3 = 10^{-4}$ for both methods ($w_2=0$ since a desired orientation is not set). It is observed that penalizing the squared norm of the velocity terms in the integrated cost with a light weight of $10^{-3}$ improves the convergence of the iLQR-VSCMO and also yields smoother motions; whereas, this is not needed for the SCvx-VSCMO owing to the trust region constraint. For a fair comparison of the cost values, the velocity component is subtracted from the cost values obtained from the iLQR-VSCMO. The initial trust region radius is $10^2$ for both applications. Both methods are initialized with a trivial initial guess in which all torque components $\boldsymbol{\tau}_u$ are zero. Each joint has a torque limit of $\pm$1 N-m. On the other hand, the virtual stiffness components are initialized and bounded above by 5~N/m, and $\alpha=15$. The time horizon is 1 s and $N=10$.

In the 7 DOF arm case, the weights of the cost function are tuned as $w_1=10^3$, $w_2=1$, and $w_3=2\times 10^{-2}$. The initial value and the upper bound of $k$ is 10~N/m and $\alpha=20$. As in the planar example, the initial $\boldsymbol{\tau}_u$ trajectories are set to zero, and the time horizon is 1 s and $N=10$.

\subsection{Software Implementation}
\subsubsection{Planar Example}
A MATLAB implementation of the control-limited iLQR is available \cite{todorov2014control}. In \cite{acikmese2018scvx}, SCvx is also implemented in MATLAB using CVX \cite{cvx}. To take advantage of these resources, we compare SCvx-VSCMO and iLQR-VSCMO methods in MATLAB. In this simulation environment, contacts are frictionless and penetrations of the links into the object and self collisions are not checked. Physical contacts of the end effector with the object are simulated through a spring model with a stiffness of $10^4$ and time steps of 2 ms, and the friction coefficient effective on the object's motion is 0.75.
\subsubsection{Standard Robot Arm}
In this case, MuJoCo \cite{mujoco} is used to model and simulate the dynamics of the 7 DOF robot with frictional contacts with the environment. The convex subproblems are solved by using the large-scale sparse solver SQOPT \cite{sqopt77}. This solver is well suited for the CITO problem since the convex subproblem has thousands of decision variables but there are only $6+N\times n_c$ non-zero elements in the quadratic cost.

For both applications, numerical differentiation (i.e., central differences) is used to approximate the derivatives. Computations are run on a workstation with Intel Core i7-6700K processor.

\section{Results}
\subsection{Comparison of SCvx-VSCMO and iLQR-VSCMO}
The two methods are compared in terms of convergence, computation time, and the quality of motions by using identical convergence conditions.
\subsubsection{Convergence Analysis}
\begin{figure}
  \centering
  \vspace{0.1cm}
  \includegraphics[width=.8\columnwidth]{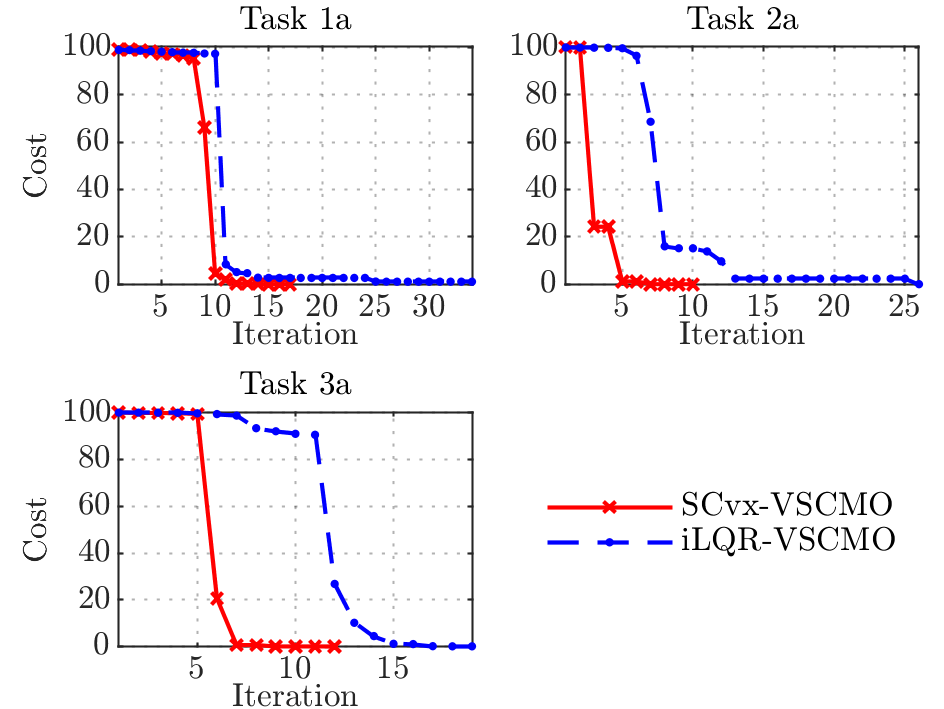}
  \caption{Change of the nonlinear cost over iterations for Tasks 1a, 2a, and 3a. The time horizon is 1 s for all cases.}
  \label{fig:convergence}
\end{figure}

The results for the convergence of the nonlinear cost for both methods are shown in Fig. \ref{fig:convergence} and yield the following observations. First, both methods find acceptable motions (i.e., with a cost value below 1) for all tasks. Second, the cost is diminished more than 90\% within 15 iterations for both methods and all cases. Third, the SCvx-VSCMO finds an acceptable motion in 12, 7, and 7 iterations; while the iLQR-VSCMO achieves that in 34, 26, and 17 iterations, respectively. Last, the SCvx-VSCMO outperforms the iLQR-VSCMO in terms of convergence speed and final cost value in all cases.

\begin{figure}
  \centering
  \includegraphics[width=.8\columnwidth]{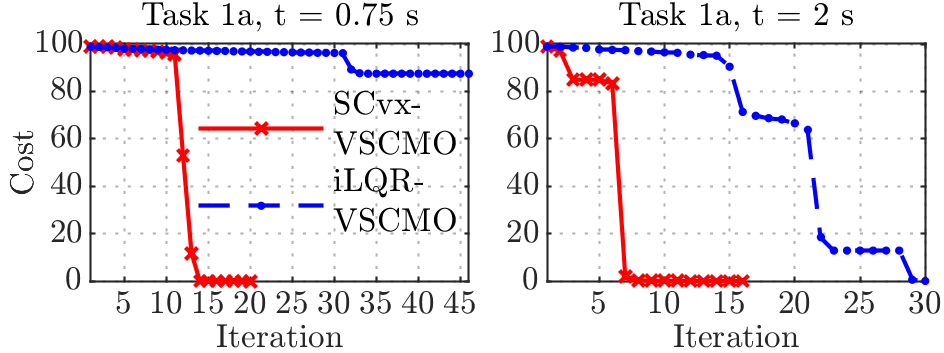}
  \caption{Change of the nonlinear cost over iterations for pushing the object 10 cm in $-x$ direction in 0.75 s and 2 s.}
  \label{fig:convergence_time}
\end{figure}

In order to see the reactions of these methods to different time horizons, we also test them for time horizons of 0.75 s and 2 s for Task 1a, i.e., pushing in $-x$ direction. Figure \ref{fig:convergence_time} demonstrates the change of the nonlinear cost for this test. We assume that the problem is smaller (i.e., involves less decision variables and constraints) but more difficult (or more complex) for the shorter time horizon since the same task needs to be done in a shorter amount of time, and larger but less complex for the longer time horizon. It is seen that the SCvx-VSCMO scales well with the size and the complexity of the task; whereas, the iLQR-VSCMO cannot find an acceptable solution for the shorter time horizon and its convergence deteriorates as the problem becomes larger.

\subsubsection{Computation Time}
Both optimal control methods rely on the first-order approximation of the nonlinear dynamics. In addition, iLQR requires first and second derivatives of the cost. On the other hand, SCvx solves a convex program in each iteration; whereas, a backward pass is performed in iLQR followed by a line search with forward passes. In order to compare the computational costs, the computation times averaged over tasks for differentiation $t_d$, solving the convex program $t_{cp}$, backward pass $t_{bp}$, and line search $t_{ls}$ per iteration as well as for an iteration $t_i$ and the total times until convergence $t_t$  are shown in Table \ref{tab:computation}.

The results show that the differentiation takes longer for the iLQR-VSCMO since it also needs the derivatives of the cost function. On the other hand, the backward pass of iLQR-VSCMO is orders of magnitude faster than SCvx-VSCMO's convex optimization since the closed-form solution of Riccati-like equations is used. However, the line search step of iLQR-VSCMO is costly because forward simulations are performed. Overall, an iteration of SCvx-VSCMO takes about 1.6 times longer than an iteration of iLQR-VSCMO. However, overall, the total time elapsed for the convergence of iLQR-VSCMO is more than twice the time it takes for SCvx-VSCMO since SCvx-VSCMO converges in fewer iterations.

\begin{table}
    \centering
    \caption{Average computation times in terms of seconds}
    \begin{tabular}{ |c|c|c|c||c|c|c|c|c| }
        \hline
        \multicolumn{4}{|c||}{SCvx-VSCMO} & \multicolumn{5}{c|}{iLQR-VSCMO}  \\
        \cline{1-9}
        $t_d$& $t_{cp}$ & $t_{i}$& $t_{t}$ & $t_d$  & $t_{bp}$ & $t_{ls}$  & $t_{i}$    & $t_{t}$  \\
        \hline
        2.20    & 0.62    & 2.87  & 37.1  & 2.46      & 0.003    & 0.32     & 1.81      & 78.1   \\
        \hline
    \end{tabular}
    \label{tab:computation}
\end{table}

\subsubsection{Quality of Motions}
The quality of motions are compared in terms of a physical inaccuracy metric $\psi$ that is defined as the integral of the norm of the virtual forces over dynamic time steps (i.e., $\psi=\int{||\boldsymbol{\gamma}(t)|| dt}$) and the positioning error that is the deviation of the final position of the object from the desired position. Table \ref{tab:quality} presents these metrics for all tasks. It is seen that the amounts of virtual forces are negligible for all cases since such small forces are already dissipated by friction. The tasks can be deemed to be completed considering the positioning errors are in millimeter level in all cases. Nonetheless, both the physical inaccuracies and the positioning errors for the SCvx-VSCMO are substantially smaller than those for the iLQR-VSCMO.

 \begin{table}
    \centering
    \caption{Physical inaccuracies and positioning errors of motions}
    \begin{tabular}{ |c||c|c||c|c| }
        \hline
        \multirow{2}{*}{Task} & \multicolumn{2}{c||}{$\psi$~[N-s]} & \multicolumn{2}{c|}{$||\mathbf{p}_o^e||$~[mm]}  \\
        \cline{2-5}
                    & SCvx      & iLQR      & SCvx      & iLQR     \\
        \hline
        1a          & 0.0001    & 0.3545    & 0.0067    & 10.2454  \\
        \hline
        2a          & 0.1545    & 0.4339    & 0.0593    & 0.7104  \\
        \hline
        3a          & 0.1741    & 1.9183    & 0.0069    & 0.5301  \\
        \hline
    \end{tabular}
    \label{tab:quality}
\end{table}

\subsection{Experiments of SCvx-VSCMO on Sawyer}
The SCvx-VSCMO is used to generate pushing behaviors for Sawyer such that the red box on the table in Fig. \ref{fig:sim_vs_exp} is moved 5 cm and 10 cm forward without rotation, i.e., Tasks 1b and 2b. The SCvx-VSCMO finds motions that pushes the box 4.62 cm and 9.26 cm with rotations of $5^{\circ}$ and $18^{\circ}$ in counter-clockwise direction about the $z$-axis, which can be deemed successful. Moreover, the optimization suppresses $k$ completely in both cases. In other words, the virtual forces vanish entirely. In order to show that the resulting motions are physically feasible, we play-back them on the hardware by using an inverse dynamics feed-forward controller. In these experiments, it is observed that the robot and the object move similarly in simulation and experiment. However, the object covers 2.5 cm and 4 cm more distance for Tasks 1b and 2b respectively. The discrepancies between simulation and experiment are likely to be caused by the errors in modeling. Nevertheless, we believe that by running this framework as a receding horizon controller such problems can be overcome.

\begin{figure}
  \centering
  \vspace{0.05cm}
  \includegraphics[width=.5\columnwidth]{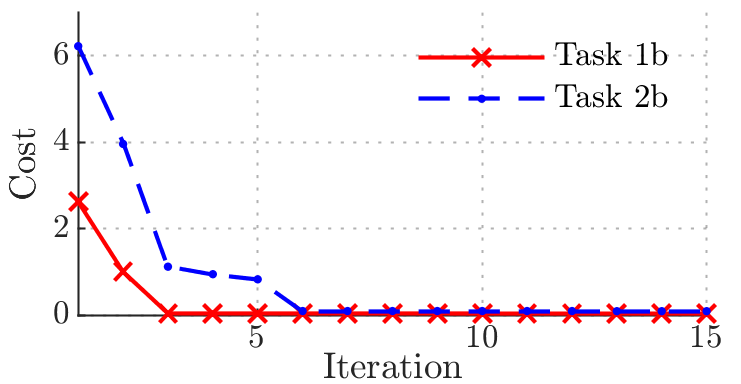}
  \caption{Change of the nonlinear cost for the SCvx-VSCMO for pushing the object 5 cm (Task 1b) and 10 cm (Task 2b) forward.}
  \label{fig:convergence_mujoco}
\end{figure}

Figure \ref{fig:convergence_mujoco} shows the convergence of the nonlinear cost for the SCvx-VSCMO for both tasks. Note that the initial final cost values are 2.5 and 10 for Tasks 1b and 2b. SCvx-VSCMO converges fast and reliably in this problem as well. The algorithm converges to an acceptable solution in 3 and 6 iterations for Tasks 1b and 2b respectively. The cost descent rate being similar for both tasks suggests that the SCvx-VSCMO scales well with the complexity of the task by assuming that pushing the object farther in the same time period is a more difficult task.

On the other hand, the average computation time for convex optimization is 60 ms, namely, an order of magnitude less than that for the planar example even though the convex problem is much larger compared to the planar example. This is achieved by exploiting the structure of the CITO problem through the large-scale and sparse solver. The convex problems may be solved even faster by using modern interior point solvers \cite{acikmese2018scvx}, such as \cite{boyd2013ecos}. In addition, numerical differentiation is both computationally expensive and prone to numeric errors. Instead, using analytic derivatives \cite{carpentier2018analytical} or auto-differentiation tools, such as \cite{buchli2017autodiff}, would speed up both convergence and computation significantly. Thus, we believe that the proposed method has the potential to run in real time in a receding horizon fashion.

\section{Conclusion}
In this study, we have presented a CITO method based on the VSCM and successive convexification (SCvx-VSCMO). We have demonstrated the proposed approach for non-prehensile manipulation applications. Moreover, the SCvx-VSCMO has been compared to an iLQR-based variant on a planar example. The results show the following. First, both methods can find physically-feasible motions that complete the tasks with a trivial initial guess owing to the VSCM. Second, the SCvx-VSCMO outperforms the iLQR-VSCMO in terms of convergence, computation time, and the quality of motions. Last, the proposed approach exhibits reliable convergence properties for different tasks and time horizons.

Additionally, the proposed framework has been tested on a standard 7 DOF robot arm. It has been shown that by utilizing a state-of-the-art physics engine and a special solver that exploits the structure of the problem, the proposed method can be used efficiently for a real-world application. Moreover, the planned motions have been executed on hardware to demonstrate that they are physically feasible. Our future work will focus on incorporating frictional forces into our contact model and improving the implementation such that the method can be used as an MPC in real time. Our results imply that the latter can be achieved by replacing numerical differentiation by analytic derivatives and employing faster convex programming solvers.

\addtolength{\textheight}{-0.12cm}   


\balance


\end{document}